\documentclass{article}

\usepackage[nonatbib, final]{nips_2016}

\usepackage[utf8]{inputenc} 
\usepackage[T1]{fontenc}    
\usepackage{hyperref}       
\usepackage{url}            
\usepackage{booktabs}       
\usepackage{amsfonts}       
\usepackage{nicefrac}       
\usepackage{microtype}      
\usepackage{subfigure}
\usepackage{graphicx}

\title{FashionNet: Personalized Outfit Recommendation with Deep Neural Network}

\author{
  Tong He\\
  University of California, Los Angeles\\
  \texttt{simpleig@cs.ucla.edu} \\
   \And
   Yang Hu \\
   University of Electronic Science and Technology of China \\
   \texttt{yanghu@uestc.edu.cn} \\
}

\begin{document}

\maketitle

\begin{abstract}
With the rapid growth of fashion-focused social networks and online shopping, intelligent fashion recommendation is now in great need. We design algorithms which automatically suggest users outfits (e.g. a shirt, together with a skirt and a pair of high-heel shoes), that fit their personal fashion preferences. Recommending sets, each of which is composed of multiple interacted items, is relatively new to recommender systems, which usually recommend individual items to users. We explore the use of deep networks for this challenging task. Our system, dubbed FashionNet, consists of two components, a feature network for feature extraction and a matching network for compatibility computation. The former is achieved through a deep convolutional network. And for the latter, we adopt a multi-layer fully-connected network structure. We design and compare three alternative architectures for FashionNet. To achieve personalized recommendation, we develop a two-stage training strategy, which uses the fine-tuning technique to transfer a general compatibility model to a model that embeds personal preference. Experiments on a large scale data set collected from a popular fashion-focused social network validate the effectiveness of the proposed networks.
\end{abstract}

\section{Introduction}

Recommender system is one of the most important tools for solving the information overload problem on the Internet. In various websites, a wide variety of stuff have been recommended to users by their recommendation engines, e.g. movies, songs, books, products, advertisements, etc. All these are objects that cannot be split further. There are scenarios where users are interested in objects that consist of multiple, interacted parts. For example, women are always obsessed about the outfits to wear tomorrow, which consist of several fashion items that should be compatible. When moving, people are puzzled by what furniture and decorations they should buy for each room of their new house. These trigger a new recommendation problem in which the stuff being recommended are object sets. In this work, we will study this set recommendation problem. Specifically, we focus on personalized outfit recommendation, which is representative and in much demand in practice.

\begin{figure}[h]
  \centering
  \begin{center}
  \subfigure[User 1]{
  \includegraphics[width=0.3\linewidth,height=1.7in]{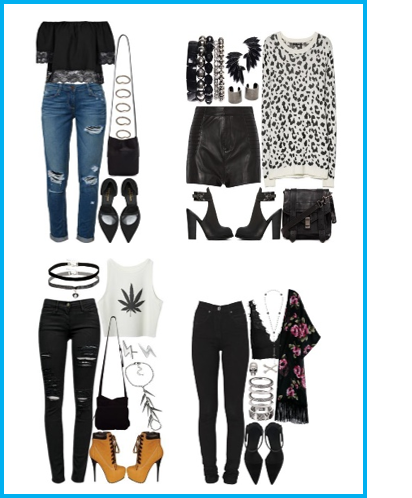}}
  \subfigure[User 2]{
  \includegraphics[width=0.34\linewidth,height=1.7in]{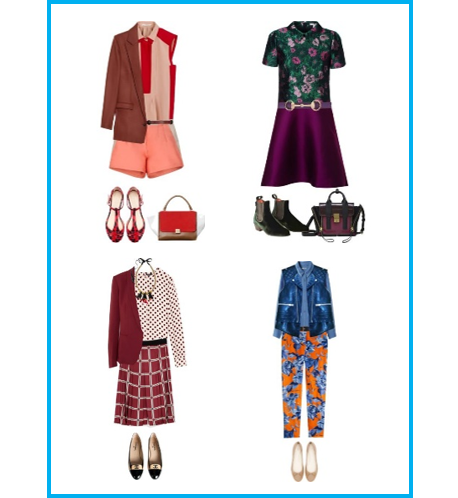}}
  \subfigure[User 3]{
  \includegraphics[width=0.3\linewidth,height=1.7in]{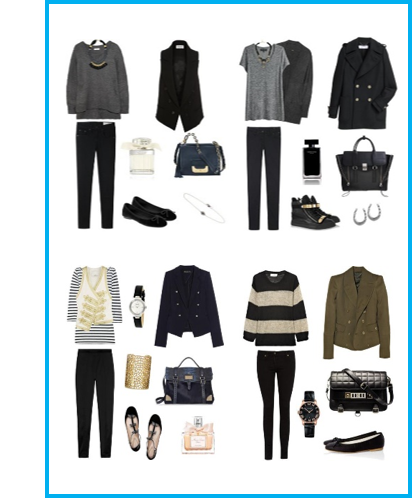}}
  \end{center}
  \caption{Examples of fashion sets created by three Polyvore users. Different users have different fashion tastes. Our goal is to recommend outfits to users that fit their style preferences.}
  \label{fig-personal}
\end{figure}

Fashion-focused online communities such as Polyvore\footnote{http://www.polyvore.com/}, Chictopia\footnote{http://www.chictopia.com/}, Lookbook\footnote{http://lookbook.nu/} have developed rapidly in recent years. People are fond of making and sharing outfits on these platforms (Figure~\ref{fig-personal} illustrates exemplar outfits created by three users on Polyvore). They also extensively browse through these websites for great outfit ideas. With the huge number of outfits shared, it is of great value to build a system that automatically recommend outfits to users that fit their personal preferences.

Due to the traditional difficulties in effectively representing the content of the items, existing recommender systems are heavily dependent on the collaborative filtering (CF) technique. CF uses the historical ratings given to items by users as the sole source of information for learning. Its performance is very sensitive to the sparsity level of the user-item matrix. In set recommendation, the ratings correspond to entries of a high-order user-item tensor. The sparsity problem is even worse than in item recommendation, which makes the CF method inapplicable directly. On the other hand, the recent progress of deep neural networks provides promising solution to the representation problem of image content~\cite{lecun1998gradient,krizhevsky2012imagenet,Chatfield14,szegedy2015going}. We therefore explore the use of deep networks for outfit recommendation. Deep representation learning for the content information and collaborative filtering for the rating information are jointly performed in this work.

There are two key problems need to be tackled in personalized outfit recommendation. The first is the modeling of the compatibility among multiple fashion items. And the second is capturing users' personal interests. Inspired by recent works on deep network based metric learning, we solve the former one by first mapping the item images to a latent semantic space with a convolutional neural network (CNN). The compatibility of the features are then measured through a multi-layer fully-connected network. We study three alternative architectures that combine feature learning and compatibility modeling in different ways. For the second problem, we encode user-specific information in the parameters of the networks. Although each user may has his unique fashion taste, he still follows some general rules for making outfits. Besides, the usually small number of training samples for a single user makes it necessary to borrow training data from other users that share similar taste. With these observations, we adopt a two-stage strategy for training the network. The first stage learns a general compatibility model from outfits of all users. In the second stage, we fine-tune the general model with user-specific data. Fine-tuning is an important technique for training deep neural networks for applications with limited number of training samples. In this work, we explore the use of fine-tuning for personalized modeling, which is a new usage of this technique. We conduct experiments with a large scale data set collected from Polyvore. Detailed comparisons between different network structures and training strategies are provided.

\section{Related work}

Driven by the huge profit potential in the fashion industry, intelligent fashion analysis, e.g. clothing recognition, parsing, retrieval, and recommendation, has received a great deal of attention recently~\cite{Liu2014Fashion}. For recommendation, existing work mainly studied it under the retrieval framework~\cite{Liu2012Hi,Jagadeesh2014Large,Iwata2011Fashion,McAuley2015Image,veit2015learning}, i.e. recommending fashion items that go well with a given query. In the latest work, McAuley et al.~\cite{McAuley2015Image} learned a parametric distance transformation such that clothing pairs that fit well are assigned with a lower distance than those that are not. Veit et al.~\cite{veit2015learning} used a Siamese CNN architecture to learn feature transformation for compatibility measure between pairs of items. All these works only studied compatibility between two objects. Besides, they only modeled general matching rules and did not consider the personalization issue. Hu et al.~\cite{hu2015collaborative} made an initial exploration of personalized outfit recommendation. A functional tensor factorization method was proposed to model the user-item and item-item interactions. However, they used hand-crafted features and did not jointly optimize feature representation and compatibility modeling.

Several recent works have explored the use of deep networks for metric learning. Some learned feature representations using CNN such that traditional metrics, e.g. the Euclidean distance, between the learned features would be consistent with the semantic distance~\cite{bell2015learning,veit2015learning,wang2014learning}. Some other works not only considered representation learning but also concatenated the features of the two objects and further learned a multi-layer network to measure the distance~\cite{Kiapour2015Where,han2015matchnet,Zbontar2015Computering}. We followed the ideas of the second kind of works. But instead of similarity, we learn compatibility of items from different categories. Researchers have also begun to apply deep learning to recommender systems. Wang et al.~\cite{Wang2015Collaborative} proposed a hierarchical Bayesian model that jointly performed deep representation learning and collaborative filtering. In~\cite{Geng2015Learning}, a deep model which learned a unified feature representation for both users and images was presented. However, these works only considered recommendation of individual items.

\section{Our approach}
\label{ourApproach}

Assume that the heterogeneous fashion items can be grouped into $N$ categories. For example, the three most essential categories for fashion are tops, bottoms and shoes. An outfit is a collection of fashion items with each coming from a different category. Given some historical data, we would like to learn a model so that for any user/outfit pair, we can assign it a rating score $s$. The score reflects the level of affection the user has for the outfit. The higher the score, the more appealing the outfit is to the user. The outfits with the highest scores are then recommended to the users.

Basically, the rating $s$ for a user/outfit pair is determined by how well the items in the outfit go with each other. It also depends on how much the user likes the items as well as the act of putting them together. We design appropriate deep network structures to model the interactions among the items. To achieve personalization, we develop a two-stage training strategy and embed the user-specific preferences in the parameters of the network. In the following, we discuss these two components in more details.

\subsection{Network architecture}
\label{models}

Our problem is closely related to metric learning, an extensively studied problem in machine learning. The metric learning task learns a distance function over objects so that the distances between pairs of objects can be effectively evaluated. In our application, we need to model relationship among multiple objects and instead of similarity, we are interested in the compatibility of objects. Nevertheless, the basic framework developed for metric learning is still applicable for solving our problem. There are mainly two steps for learning a distance metric, i.e. mapping the objects to a low dimensional latent space and measuring the distance in that space. We design components functionally similar in our networks.

\begin{figure}[t]
\begin{center}
\subfigure[\small{FashionNet A}]{
\includegraphics[height=1.05in,width=0.85in]{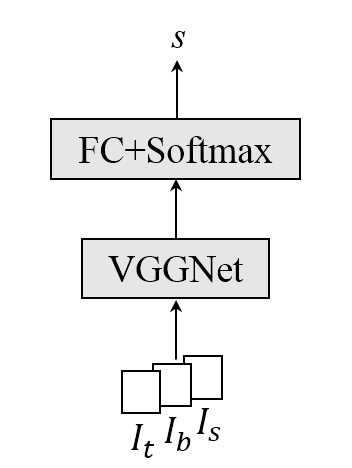}}
\subfigure[FashionNet B]{
\includegraphics[height=1.05in]{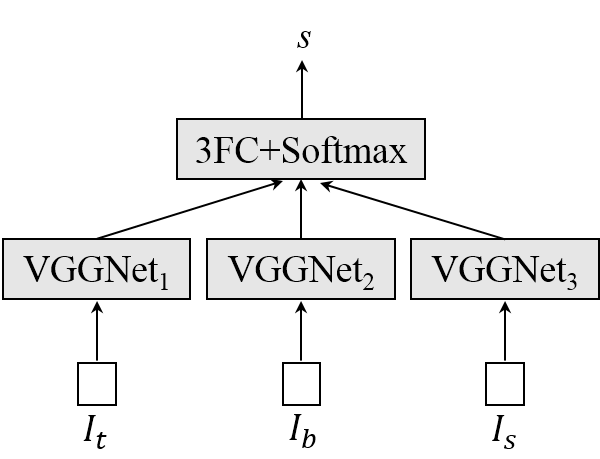}}
\subfigure[FashionNet C]{
\includegraphics[height=1.25in]{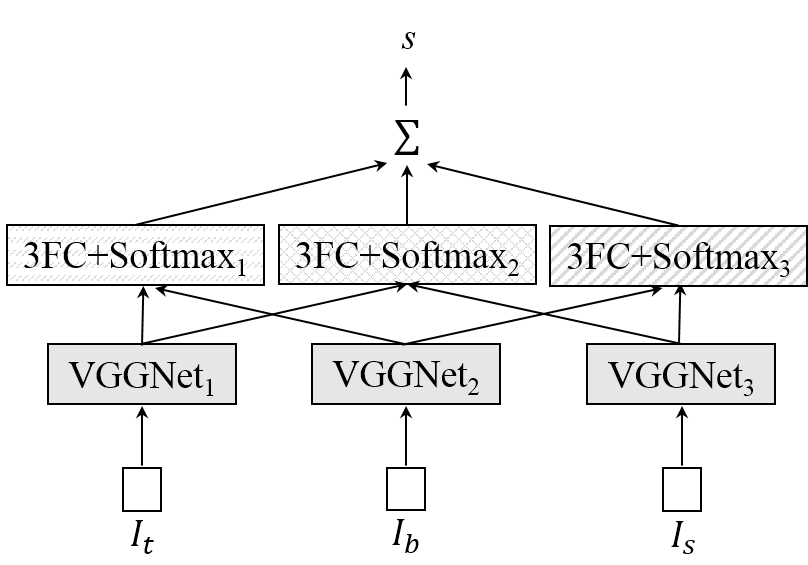}}
\subfigure[Training structure]{
\includegraphics[height=1.05in]{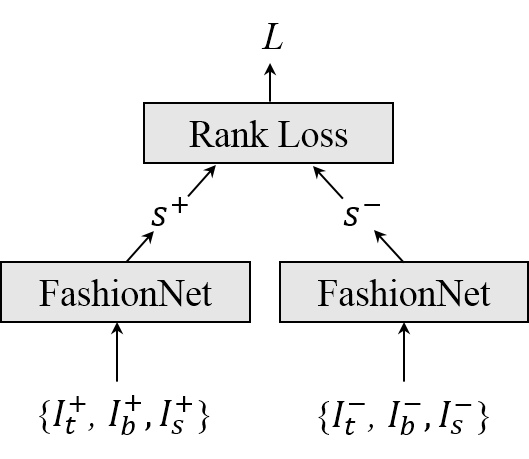}
\label{fig-rankloss}}
\end{center}
\caption{Network architectures. (1), (2) and (3) are three alternative architectures. For VGGNet, we use the VGG-Medium structure in~\cite{Chatfield14}. FC stands for fully-connected layer. For both VGGNet and the FC layers, we use Local Response Normalization and ReLU activation. $I_t$, $I_b$ and $I_s$ are images of the top, bottom and shoes respectively. Images are preprocessed before fed into the network. The preprocessing includes two steps, i.e. spatially resizing to $224\times224$ and subtracting the mean images. $s$ is the predicted preference score of the input. (4) illustrates the two-tower structure for training. The two towers share the same set of parameters.}
\label{fig:structure}
\end{figure}

We explore three different network architectures as shown in Figure~\ref{fig:structure} and Table~\ref{tab-models}. Without loss of generality, here we assume an outfit consists of three items, i.e. a top, a bottom and a pair of shoes. It is straight forward to extend the networks for handling outfits with more items.

1) In FashionNet A, the images of the items are first concatenated to create a new image with 9 color channels. Then the compounded image is fed to a widely used CNN model, e.g. the VGGNet. The output layer is a fully-connected layer with the Softmax function as its activation function. It converts the feature representation obtained by the CNN model to some scores. There are two outputs from this final layer. They can be interpreted as estimates of the probability that the outfit is liked and disliked by the user, respectively. In this architecture, the components of representation learning and compatibility measure are fully integrated. The two steps are carried out simultaneously right from the first convolution layer.

2) In FashionNet B, we apply representation learning and compatibility measure sequentially. The images are first mapped to a feature representation via a feature network. The same CNN model is used for items from different categories, i.e. item images are projected to a common latent space. To model the compatibility, we concatenate the features of all items and feed them to three fully-connected layers. Some recent works have used this type of  fully-connected network for distance metric learning, with applications in patch-based matching~\cite{han2015matchnet}, cross-domain retrieval~\cite{Kiapour2015Where} and stereo matching~\cite{Zbontar2015Computering}. In this work, we show that this network structure also has the capacity for approximating the underlying compatibility among multiple features.

3) Both FashionNet A and B try to directly model the compatibility among multiple items. They may come across some difficulties when trying to capture the high-order relationships. The data space is significantly expanded when we concatenate all the items, either in original image space as FashionNet A or in feature space as FashionNet B. Due to the curse of dimensionality, a huge number of training samples may be required for a good model to be learned. Although users on the Internet have contributed so many outfit ideas, it is still minor compared to the number of all possible outfits.
To alleviate this problem, we impose a prior restraint in FashionNet C. We assume that the compatibility of a set of items is mainly determined by how well pairs of these items go with each other. Therefore, after representation learning, we only concatenate features of any two items and load them into a matching network. We again use a three-layer fully-connected network with Softmax activation at the third layer to model the pairwise compatibility. Different matching networks are used for different types of pairs, e.g. the matching between tops and bottoms is modeled differently from the matching between tops and shoes. Then the outputs from the final layers regarding the probabilities that the item pairs are matched well are added together to get a final score $s$ for the whole outfit.

We formulate this learning task as a learning-to-rank problem. Unlike binary classification, which only models whether an outfit is liked or disliked by a user, the rank learning formulation can handle multiple levels of ratings. We use pairs of outfits as the training samples. A training sample contains two outfits, e.g. $\{I_t^+,I_b^+,I_s^+\}$ and $\{I_t^-,I_b^-,I_s^-\}$, where the former is preferable to the latter. We use a two-tower structure illustrated in Figure~\ref{fig-rankloss} to train the networks. Specifically, we minimize the rank loss
\begin{equation}
L = \frac{1}{M}\sum_{i=1}^M\log(1+\exp(-(s_i^+-s_i^-)))
\end{equation}
over a training set. $s_i^+$ and $s_i^-$ are the predicted scores for the outfits in the $i$-th pair.

\begin{table}
  \caption{Layer parameters of FashionNets.}
  \label{tab-models}
  \centering
    \begin{tabular}{lllll}
    \toprule
    Architecture   & Layers              & Shared parameters & Input size & Output size \\
    \midrule
    FashionNet A & VGGNet & No & 9x224x224 & 2048 \\
    ~            & FC + Softmax & No & 2048 & 2 \\
    \midrule
    ~            & VGGNet$_1$ & \textbackslash & 3x224x224 & 2048 \\
    FashionNet B & VGGNet$_2$ & \ | Yes & 3x224x224 & 2048 \\
    ~            & VGGNet$_3$ & / & 3x224x224 & 2048 \\
    ~            & 3FC + Softmax & No & 6144 & 2 \\
    \midrule
    ~            & VGGNet$_1$ & \textbackslash & 3x224x224 & 2048 \\
    ~            & VGGNet$_2$ & \ | Yes & 3x224x224 & 2048 \\
    FashionNet C & VGGNet$_3$ & / & 3x224x224 & 2048 \\
    ~            & 3FC + Softmax$_1$ & No & 4096 & 2 \\
    ~            & 3FC + Softmax$_2$ & No & 4096 & 2 \\
    ~            & 3FC + Softmax$_3$ & No & 4096 & 2 \\
    \bottomrule
    \end{tabular}
\end{table}

\subsection{Training}

For a single user, we usually only have a small number of training outfits. It is difficult to directly train a network from scratch with such limited training data. On the other hand, there are a large number of outfits shared by other users who have similar fashion taste. It would be helpful to take advantage of this data. Furthermore, although each user may have his own preference, there are some customary rules followed by most people for making outfits, e.g. T-shirt and jeans are usually paired up. With these observations, we design a two-stage procedure to train the deep network for personalized outfit recommendation.

In the first stage, we learn a general model for compatibility measure. We discard the user information and mix outfits created by different users together. Following~\cite{hu2015collaborative}, we create neutral outfits by mixing randomly selected fashion items. It is reasonable to assume that the items in a user created outfit is more compatible than those in a neutral outfit. Therefore a training sample can be made by pairing a user generated outfit with a neutral one. We initialize the VGGNet with parameters pretrained on ImageNet~\cite{deng2009imagenet} and initialize the other layers with random numbers drawn from Gaussian distribution. Then we optimize the parameters of the whole networks using the mixed dataset.

In the second stage, we train user-specific model for personalized recommendation. For each user, we first initialize the network with parameters obtained by the previous general training. Then we use each user's own data to fine-tune the parameters. Fine-tuning is a very import technique for training deep networks. It helps ease the data insufficiency problem in many applications. Here we use fine-tuning to adapt a general compatibility model to a user-specific model that addresses user's unique preference.

For FashionNet A, we fine-tune the whole network in this stage. For FashionNet B and C, there are two strategies. The first is to fine-tune the whole network so that both the feature network and the matching network will have personalized parameters. This results in different feature representations of the items for different users. The second method is to freeze the feature network and only fine-tune the matching network. In this case, the features will keep the same and the user-specific information will be carried only by the matching network. This will save a lot of computation during testing which is favorable in practice. We will compare the performance of these two strategies in Section~\ref{comparisons}.

\section{Experimental results}

\begin{table}
\caption{Results comparison of different architectures. \emph{Initial} is the result of using initial parameters. \emph{Stage one} is the result of the general model, i.e. the result after the first stage training. \emph{Stage two (partial) and stage two (whole)} are the results after the second stage of fine-tuning. \emph{Stage two (partial)} corresponds to the strategy of freezing the feature network and only updating the matching network. \emph{Stage two (whole)} fine-tunes the whole network. \emph{Stage two (direct)} is the result of directly fine-tune the initial parameters without the first stage of general training.}
\label{tab-ndcg}
\centering
\begin{tabular}{clcc}
\toprule
Architecture & Training strategy & Mean NDCG & Positive outfits in top 10 results \\
\midrule
    ~             & Initial                         & 0.35356 & 1.51375 \\
    FashionNet A  & Stage one                       & 0.59291 & 4.68125 \\
    ~             & Stage two (direct)                & 0.61545 & 5.30125 \\
    ~             & Stage two (whole)               & 0.70696 & 6.60375 \\
    \midrule
    ~             & Initial                         & 0.35552 & 1.56250 \\
    FashionNet B  & Stage one                       & 0.65694 & 5.31625 \\
    ~             & Stage two (partial)             & 0.75818 & 7.12250 \\
    ~             & Stage two (whole)               & 0.79644 & 7.83000 \\
    \midrule
    ~             & Initial                         & 0.32152 & 1.22250 \\
    FashionNet C  & Stage one                       & 0.66534 & 5.60625 \\
    ~             & Stage two (partial)             & 0.78378 & 7.60500 \\
    ~             & Stage two (whole)               & 0.81208 & 8.18250 \\
    \bottomrule
\end{tabular}
\end{table}

\subsection{Data set and evaluation criteria}

To train the deep networks, we collect data from the popular fashion website Polyvore. Our data set contains outfits created by 800 users. This is significantly larger than the data set used by previous work~\cite{hu2015collaborative}, which only contains outfits of 150 users. We refer to user created outfit as positive outfit. We also created a set of neutral outfits for each user by randomly mixing the tops, bottoms and shoes. For each user, we split the user created outfits and neutral outfits into three groups, i.e. the training, validation, and testing sets. The numbers of positive outfits per user for each group are 202, 46, and 62 respectively. The numbers for the neutral outfit are 6 times larger than the corresponding positive ones, i.e. 1212, 276, and 372.

During evaluation, for each user, the positive and neutral outfits in the testing set are ranked in descending order of their scores $s$. We use NDCG, a widely used criteria for comparing ranked lists, to evaluate the performance. The NDCG at the $m$-th position of an ordering $\pi^{'}$ is
\begin{equation}
NDCG@m = (N_{m})^{-1}\sum_{i=1}^{m}\frac{(2^{y_{\pi^{'}(i)}}-1)}{log_{2}(max(2,i))},
\label{equ-ndcg}
\end{equation}
where $N_{m}$ is the score of an ideal ranking, $y_{\pi^{'}(i)}$ is 1 for positive outfits and 0 for neutral ones. The optimal value of NDCG is 1. Mean NDCG is the mean of NDCG@m for $m = 1, \ldots, M$ with $M$ being the length of the ordering. We report the average of mean NDCG and the average of NDCG@m for some $m$ over all users. We also count the number of positive outfits in the top-k results for some $k$ and report the average number over all users.

\subsection{Results}
\label{comparisons}

\begin{figure}[h]
  \centering
  \begin{center}
  \subfigure[FashionNet A]{
  \includegraphics[width=0.32\linewidth, height=1.6in]{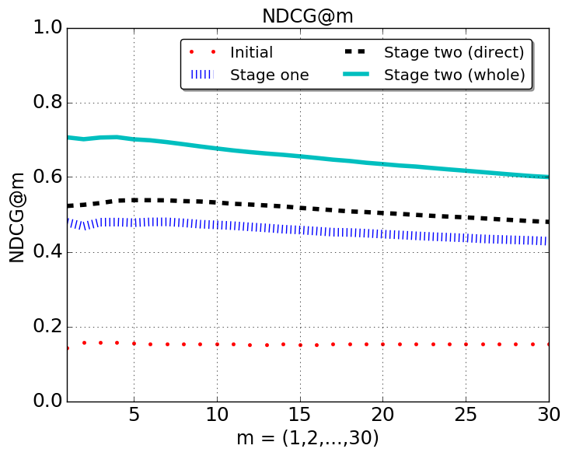}}
  \subfigure[FashionNet B]{
  \includegraphics[width=0.32\linewidth, height=1.6in]{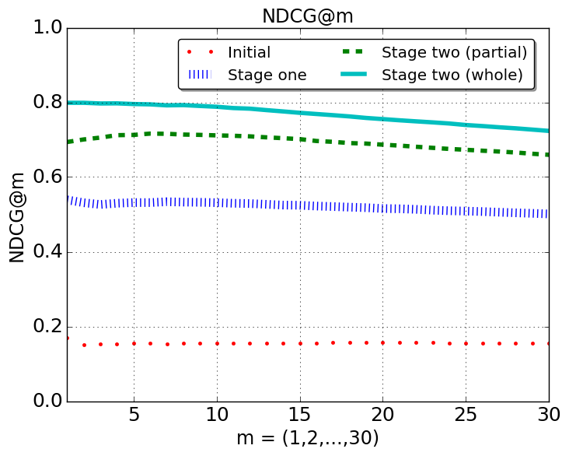}}
  \subfigure[FashionNet C]{
  \includegraphics[width=0.32\linewidth, height=1.6in]{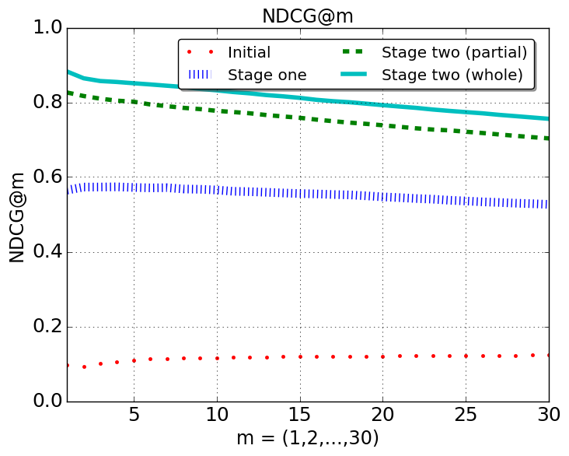}}
  \end{center}
  \caption{The average of NDCG@m over all users.}
  \label{fig-ndcgat}
\end{figure}

\begin{figure}[h]
  \centering
  \begin{center}
  \subfigure[FashionNet A]{
  \includegraphics[width=0.32\linewidth, height=1.6in]{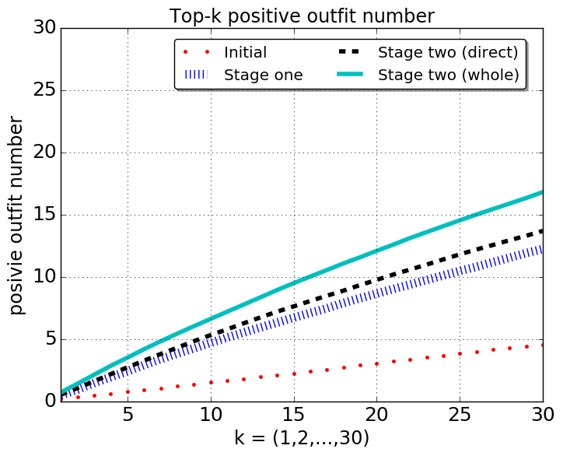}}
  \subfigure[FashionNet B]{
  \includegraphics[width=0.32\linewidth, height=1.6in]{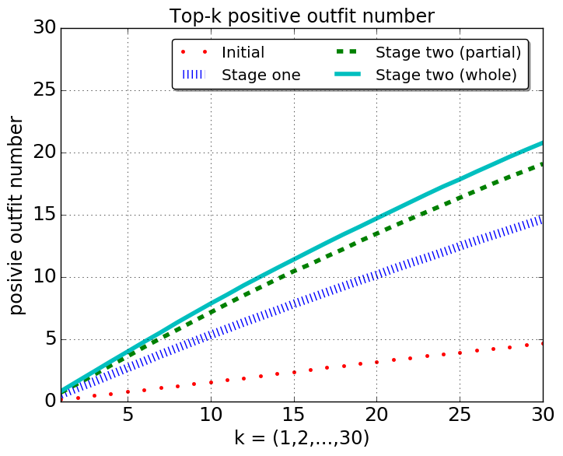}}
  \subfigure[FashionNet C]{
  \includegraphics[width=0.32\linewidth, height=1.6in]{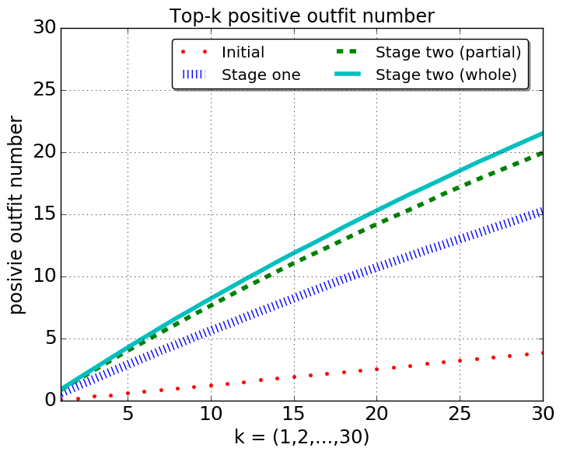}}
  \end{center}
  \caption{The positive outfit numbers in top-k results.}
  \label{fig-topkaccuracy}
\end{figure}

\begin{figure}[h]
  \begin{center}
  \subfigure[User 1]{
  \includegraphics[width=0.45\linewidth, height = 2.8in]{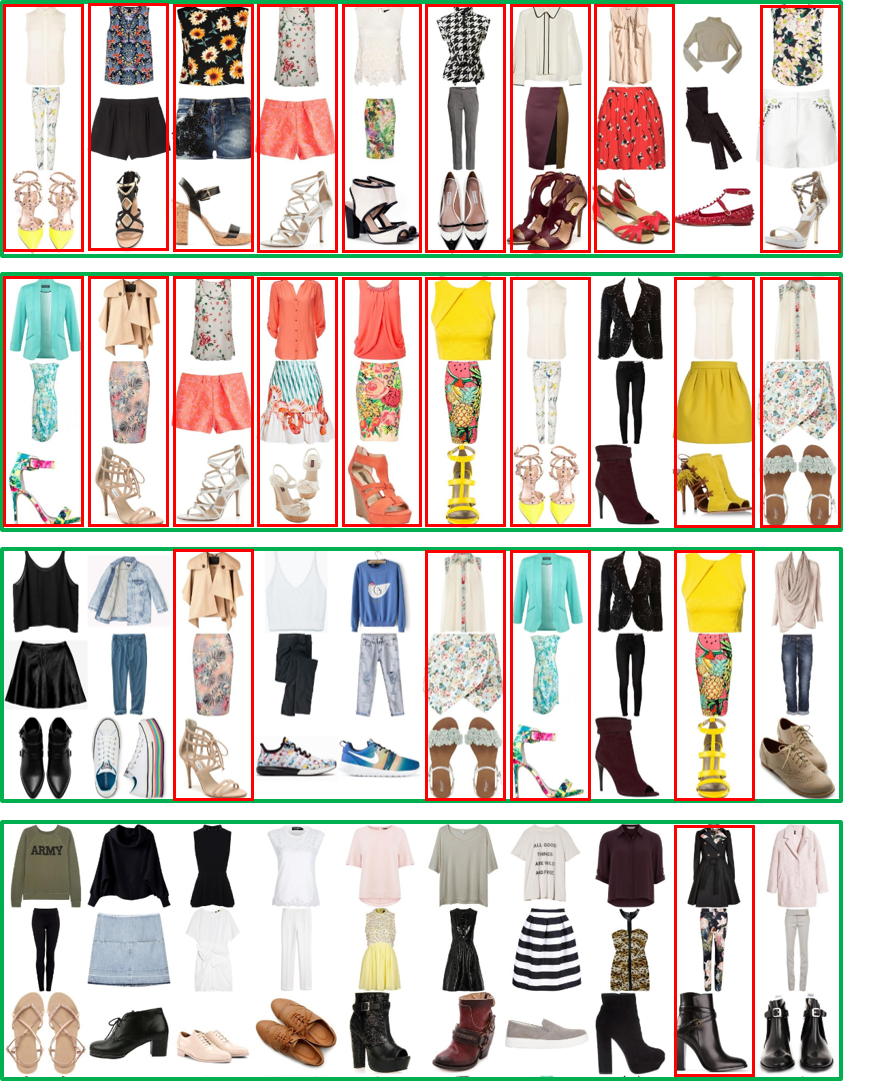}}
  \subfigure[User 2]{
  \includegraphics[width=0.45\linewidth, height = 2.8in]{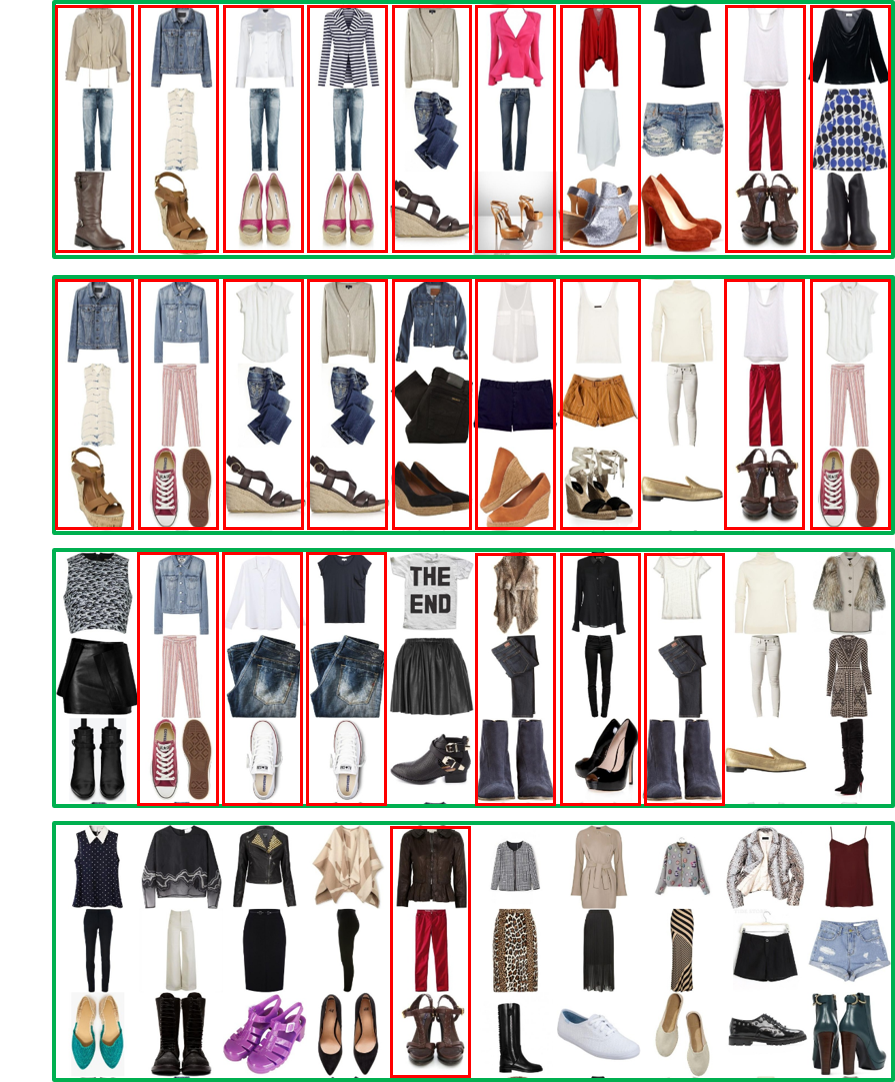}}
  \subfigure[User 3]{
  \includegraphics[width=0.45\linewidth, height = 2.8in]{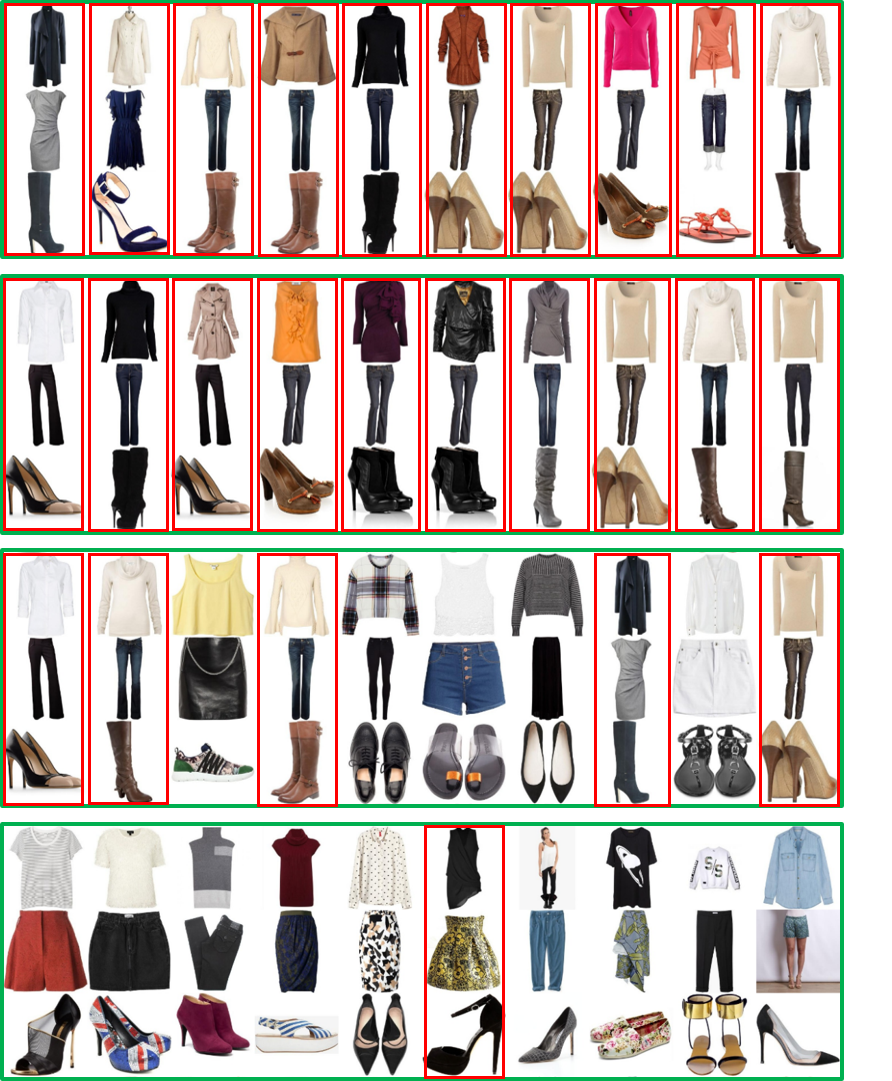}}
  \subfigure[User 4]{
  \includegraphics[width=0.45\linewidth, height = 2.8in]{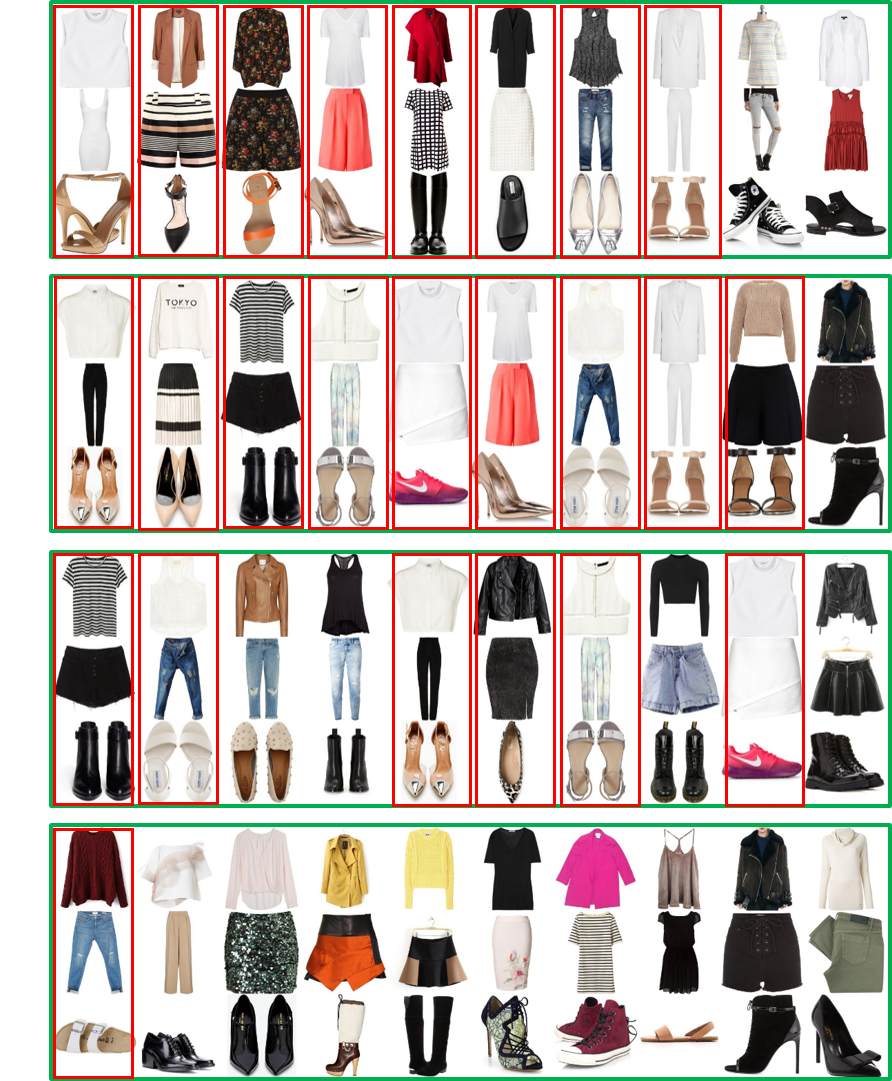}}
  \end{center}
  \caption{For each user, the top 10 outfits obtained after different training steps for FashionNet C are shown respectively. Outfits in red box are user created ones. For each user, the four results from top to bottom are: Stage two (whole), Stage two (partial), Stage one, and Initial respectively.}
  \label{fig-outfits}
\end{figure}

We use Caffe~\cite{jia2014caffe} to build the networks. For all training strategies, we set the batch size as 30 and the epoch number as 18. Learning rates for different layers during fine-tuning are determined as suggested in~\cite{Chatfield14}. For each architecture, we report the results after different training steps in Table~\ref{tab-ndcg}. The first is the result using the initial parameters, i.e. the parameters pretrained on ImageNet for the VGGNet and random values for the fully-connected layers. Then the results after the first and the second stage of training are showed. For the second stage, we compare the results of only fine-tuning the matching network and fine-tuning the entire network for FashionNet B and C. For FashionNet A, since representation learning and matching modeling are inseparable, we are not able to report the result of only fine-tuning the matching network. Instead, we check the result of passing over the first stage of general training and directly fine-tuning the network with user-specific data. In Figure~\ref{fig-ndcgat} and Figure~\ref{fig-topkaccuracy}, we show the average of NDCG@m and the number of user created outfits in top-k results respectively. Also, the top 10 outfits after each step for four users are showed in Figure~\ref{fig-outfits} for qualitative comparison.

From the experimental results, we have the following observations.

\paragraph{\textbf{FashionNet A vs. FashionNet B and C}} The performance of FashionNet A is inferior to the other two architectures. The ''stage two (whole)'' performance of FashionNet B surpasses A by about $12.7\%$ in mean NDCG and $18.6\%$ in top-10 positive outfit number. As for FashionNet C, the two margins are $14.9\%$ and $23.9\%$ respectively. The possible reason for FashionNet B and C to obtain the advantage is, representation learning and compatibility modeling are performed separately in them, so that we are able to use different network structures to achieve different functionalities. The networks are more easier to design and optimize in this case.

\paragraph{\textbf{FashionNet B vs. FashionNet C}} FashionNet C works slightly better than FashionNet B. This verified our analysis in Section~\ref{models}. By concatenating features of all the items in FashionNet B, the data space is significantly expanded, which brings in difficulties for learning. It is challenging to directly model the joint compatibility among multiple fashion items. Decomposing the high-order relationship into a set of pairwise interactions is an effective solution.

\paragraph{\textbf{Stage one vs. Stage two}} For all three architectures, there are obvious performance promotions when moving from stage one to stage two. This demonstrates the successfulness of using the fine-tuning technique for personalized modeling. The performance gaps between the two also verify the importance of personalized modeling, which has been left out by some previous works. For FashionNet A, the result of directly fine-tuning the network without general training is much worse than the two-stage training method. Therefore, it is indeed valuable to exploiting the information shared by other people when performing personalized recommendation.

\paragraph{\textbf{Stage two (partial) vs. Stage two (whole)}} For both FashionNet B and FashionNet C, fine-tuning the entire network achieves better results than only fine-tuning the matching network in the second stage of training. It indicates that learning a user-specific feature representation is helpful for the recommendation task. However, in this case we need to re-compute the features of the items for each user, which is computationally expensive. By only fine-tuning the matching network, we use the same feature representation for all users. For different users, we only need to feed the pre-computed features into the corresponding matching network. This is efficient both for training and testing. In practice, people may prefer the partial strategy although its performance is slightly worse.

\section{Conclusion}
\label{conclusions}

In this work, we have studied the problem of personalized outfit recommendation. This is a representative application of object set recommendation, a relatively new setting for recommender systems. We explored the use of deep neural networks, which jointly perform representation learning and compatibility modeling, for this task. Three alternative network architectures that combine the two functional components in different ways are designed and compared. To achieve personalized recommendation, we developed a two-stage training strategy. Fine-tuning is used to adapt a general matching model to a model that embeds user's unique fashion taste. We conducted extensive experiments on a large scale real world data set. The effectiveness of our networks has been verified by the results.

{\small
\bibliographystyle{ieeetr}
\bibliography{nips_2016}
}

\end{document}